\documentclass[lettersize,journal]{IEEEtran}
\usepackage{amsmath,amsfonts}
\usepackage{algorithmic}
\usepackage{algorithm}
\usepackage{array}
\usepackage[caption=false,font=normalsize,labelfont=sf,textfont=sf]{subfig}
\usepackage{textcomp}
\usepackage{stfloats}
\usepackage{url}
\usepackage{verbatim}
\usepackage{graphicx}
\usepackage{cite}
\usepackage{caption}
\usepackage{booktabs}   
\usepackage[table]{xcolor} 
\usepackage{graphicx}  
\usepackage{multirow}   
\usepackage{orcidlink}
\usepackage{placeins}
\usepackage{cuted}
\usepackage{capt-of}
\begin{document}

\title{RayMamba: Ray-Aligned Serialization for Long-Range 3D Object Detection}

\author{Cheng~Lu\,\orcidlink{0009-0006-5324-5361},
        Mingqian~Ji,
        Shanshan~Zhang\,\orcidlink{0000-0003-4013-6300},
        Zhihao~Li\,\orcidlink{0000-0002-0316-994X},
        and Jian~Yang\,\orcidlink{0000-0003-4800-832X}%
\thanks{Cheng Lu, Mingqian Ji, Shanshan Zhang, Zhihao Li, and Jian Yang are with the School of Computer Science and Technology, Nanjing University of Science and Technology, Nanjing 210094, China.}%
\thanks{Shanshan Zhang is the corresponding author.}%
\thanks{E-mail: lucheng@njust.edu.cn; mingqianji1299@gmail.com; shanshan.zhang@njust.edu.cn; zhihaoli828@njust.edu.cn; csjyang@njust.edu.cn.}%
}

\maketitle

\begin{abstract}
Long-range 3D object detection remains challenging because LiDAR observations become highly sparse and fragmented in the far field, making reliable context modeling difficult for existing detectors. To address this issue, recent state space model (SSM)-based methods have improved long-range modeling efficiency. However, their effectiveness is still limited by generic serialization strategies that fail to preserve meaningful contextual neighborhoods in sparse scenes. To address this issue, we propose RayMamba, a geometry-aware plug-and-play enhancement for voxel-based 3D detectors. RayMamba organizes sparse voxels into sector-wise ordered sequences through a ray-aligned serialization strategy, which preserves directional continuity and occlusion-related context for subsequent Mamba-based modeling. It is compatible with both LiDAR-only and multimodal detectors, while introducing only modest overhead. Extensive experiments on nuScenes and Argoverse 2 demonstrate consistent improvements across strong baselines. In particular, RayMamba achieves up to 2.49 mAP and 1.59 NDS gain in the challenging 40--50 m range on nuScenes, and further improves VoxelNeXt on Argoverse 2 from 30.3 to 31.2 mAP.
\end{abstract}
\begin{IEEEkeywords}
3D Object Detection, Long-range Perception, State Space Models.
\end{IEEEkeywords}

\section{Introduction}

\begin{figure}[t]
    \centering
    \includegraphics[width=\linewidth]{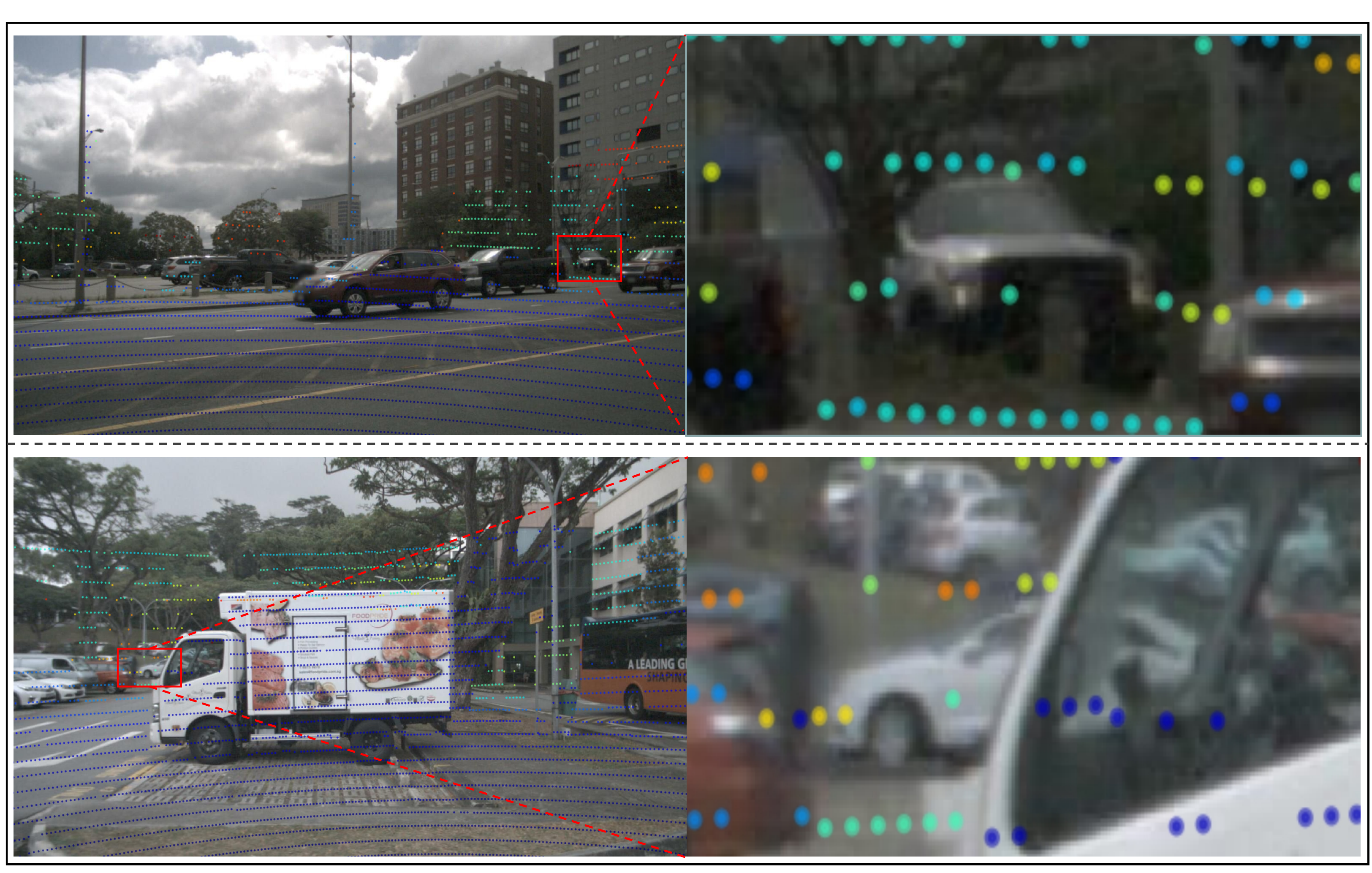}
    \caption{\textbf{Due to occlusion and distance-induced sparsity in LiDAR, distant objects are often represented by only a few returns. }}
    \label{fig:raymamba_compare}
    \vspace{-1em}
\end{figure}

\IEEEPARstart{3}{D} object detection plays a central role in autonomous driving, as it provides the spatial understanding needed to localize and classify surrounding objects. In practical driving scenarios, especially at high speeds or in complex traffic environments, reliable detection of distant objects is crucial for safe motion planning. However, long-range perception remains a major challenge for LiDAR-based detectors. On the nuScenes dataset\cite{nuscenes}, which uses a 32-beam LiDAR sensor, objects beyond roughly 40 meters are typically represented by fewer than ten returns due to distance-induced sparsity and foreground occlusion, as illustrated in Fig.~\ref{fig:raymamba_compare}. Argoverse 2\cite{argoverse}, which uses dual 32-beam LiDARs, exhibits a similar high-sparsity issue beyond roughly 50 meters. This severe degradation makes accurate long-range 3D detection substantially more difficult than near-range and mid-range perception.

This difficulty also poses a challenge to existing detector architectures. Earlier methods, such as sparse convolution-based detectors~\cite{second,voxelnext} and window-based Transformer variants~\cite{sphericaltransformer,transformer1,transformer2,transformer3}, are largely constrained by local interactions and therefore struggle to capture reliable context in sparse far-field scenes. Although recent sparse attention designs~\cite{transformer4,transformer5} and Mamba-based\cite{unimamba,voxelmamba,mambafusion} backbones enable more efficient long-range modeling, their effectiveness remains limited when the underlying geometric structure is already severely sparse and fragmented. In such cases, the key challenge is no longer simply how to enlarge the receptive field, but whether the model can establish useful contextual relationships from incomplete observations.
\begin{figure*}[t]
    \centering
    \includegraphics[width=\textwidth]{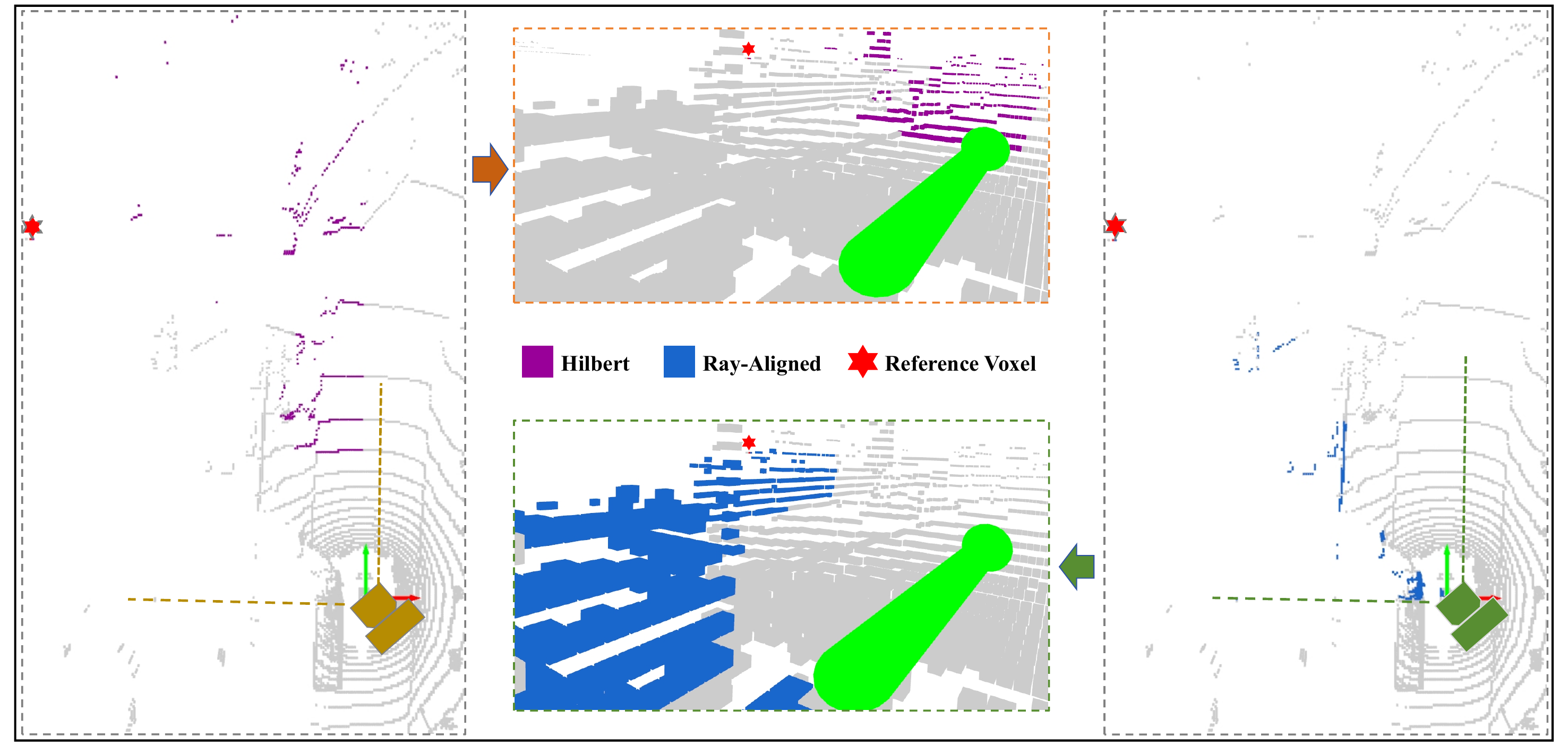}
    \caption{\textbf{Comparison of 1D sequence context in long-range sparse scenes.} For a given far-field reference voxel (red star), we highlight its context window of $K=360$ adjacent voxels in the serialized sequence. Our ray-aligned ordering (blue) preserves directionally coherent physical structures, whereas the Hilbert ordering (purple) activates spatially scattered, unrelated regions.}
    \label{fig:serialization_compare}
\end{figure*}

To address this challenge, we propose \textbf{RayMamba}, a geometry-aware state space module tailored for long-range 3D object detection. Instead of relying on generic ordering rules such as Hilbert curves, Z-order curves, or coordinate-based sorting, RayMamba adopts a ray-aligned serialization strategy that is better matched to sparse far-field scenes. Specifically, voxels are partitioned into azimuth sectors and ordered within each sector according to vertical and angular continuity, forming a sector-wise layered angular ordering. This sector-wise ordering reduces irrelevant background mixing while preserving directional and occlusion-related context. As shown in Fig.~\ref{fig:serialization_compare}, the serialized context of a far-field reference voxel differs substantially under different ordering strategies. Hilbert ordering introduces many spatially scattered, irrelevant activated voxels in the context window under highly sparse long-range conditions. By contrast, our ray-aligned ordering preserves more meaningful context, especially foreground occlusion cues with directional consistency.

The ray-aligned ordering rule of RayMamba is precomputed offline as a dense sector template, while runtime sequence construction is performed by CUDA-based parallel sorting over active voxels, making it an efficient plug-and-play module for existing sparse convolutional backbones. We primarily evaluate RayMamba on the nuScenes benchmark with CenterPoint\cite{centerpoint} and MV2DFusion\cite{mv2dfusion}, representing LiDAR-only and multimodal 3D detection frameworks, respectively. In addition, to further examine its generalization under more extreme long-range sparse conditions, we validate RayMamba with VoxelNeXt\cite{voxelnext} on the Argoverse 2 dataset. Extensive experiments on nuScenes and Argoverse 2 demonstrate that RayMamba consistently improves both LiDAR-only and multimodal detectors, with particularly notable gains in the long-range regime.

Our contributions are threefold:
\begin{itemize}
    \item We identify that, in long-range sparse scenes, a key limitation of existing Mamba-based 3D detectors lies in generic serialization, which often fails to preserve useful contextual neighborhoods during sequence modeling.
    
    \item We propose RayMamba, a ray-aligned sector-wise sequence modeling module that reduces irrelevant background mixing while preserving directional and occlusion-related context for sparse voxel modeling.
    
    \item Extensive experiments on nuScenes and Argoverse 2 verify RayMamba as a lightweight and effective plug-and-play enhancement across LiDAR-only and multimodal detectors. It yields consistent overall gains on strong baselines, and more importantly, achieves up to 2.49 mAP / 1.59 NDS improvement in the challenging 40--50 m range on nuScenes, while further improving VoxelNeXt on Argoverse 2 from 30.3 to 31.2 mAP.
\end{itemize}

\section{Related Work}

\subsection{LiDAR-based 3D Object Detection}
LiDAR-based 3D object detectors mainly include point-based~\cite{point1,point2,point3,point4,point5} and voxel-based methods~\cite{voxelnet,voxel1,voxelnext,voxel3,centerpoint,focalconv,largekernel3d,link}. In large-scale driving scenes, voxel-based methods have become the dominant paradigm due to their efficient feature extraction with voxel encoders and sparse convolutional backbones.

However, long-range detection remains challenging. Many detectors rely on dense BEV prediction, whose cost increases rapidly when higher spatial resolution is needed to preserve distant object details. Fully sparse pipelines such as VoxelNeXt~\cite{voxelnext} and FSD~\cite{fsd} have therefore been explored to avoid expensive dense BEV representations. Meanwhile, LiDAR observations become increasingly sparse and incomplete with distance due to reduced sampling density and foreground occlusion. Prior works, such as FocalConv~\cite{focalconv}, Large Kernel Conv~\cite{largekernel3d}, and LINK~\cite{link}, attempt to alleviate the resulting context loss by enlarging sparse interactions or receptive fields. Nevertheless, when distant observations are already highly sparse and fragmented, enlarging the receptive field alone is often insufficient, motivating sequence-based alternatives for long-range context modeling.

\subsection{State Space Models and Serialization in 3D Perception}
State space models (SSMs), especially Mamba-style architectures, have recently attracted growing attention in computer vision because of their linear complexity and scalable long-range sequence modeling ability. PointMamba~\cite{pointmamba} extends Mamba to 3D point cloud analysis. In 3D detection, VoxelMamba~\cite{voxelmamba} employs group-free voxel serialization with dual-scale SSMs for global context modeling, LION~\cite{lion} introduces linear recurrent modeling with window- and axis-based voxel serialization, and UniMamba~\cite{unimamba} combines sparse 3D convolution with bidirectional SSMs using complementary Z-order serialization and local-global sequential aggregation.

More generally, point cloud serialization converts unordered 3D points or voxels into 1D sequences for sequence modeling. Existing strategies include space-filling curves such as Z-order and Hilbert, as adopted in PointMamba~\cite{pointmamba}, UniMamba~\cite{unimamba}, MambaFusion~\cite{mambafusion}, and VoxelMamba~\cite{voxelmamba}, as well as window partitioning and coordinate-axis sorting, as in SWFormer~\cite{swformer}, DSVT~\cite{dsvt}, PCM~\cite{pcm}, and LION~\cite{lion}. However, these generic serialization strategies usually do not explicitly consider LiDAR acquisition geometry.

This limitation becomes more pronounced in sparse far-field scenes, motivating sensor-geometry-aware serialization for long-range 3D detection.

\section{Methodology}

\begin{figure*}[t]
    \centering
    \includegraphics[width=\textwidth]{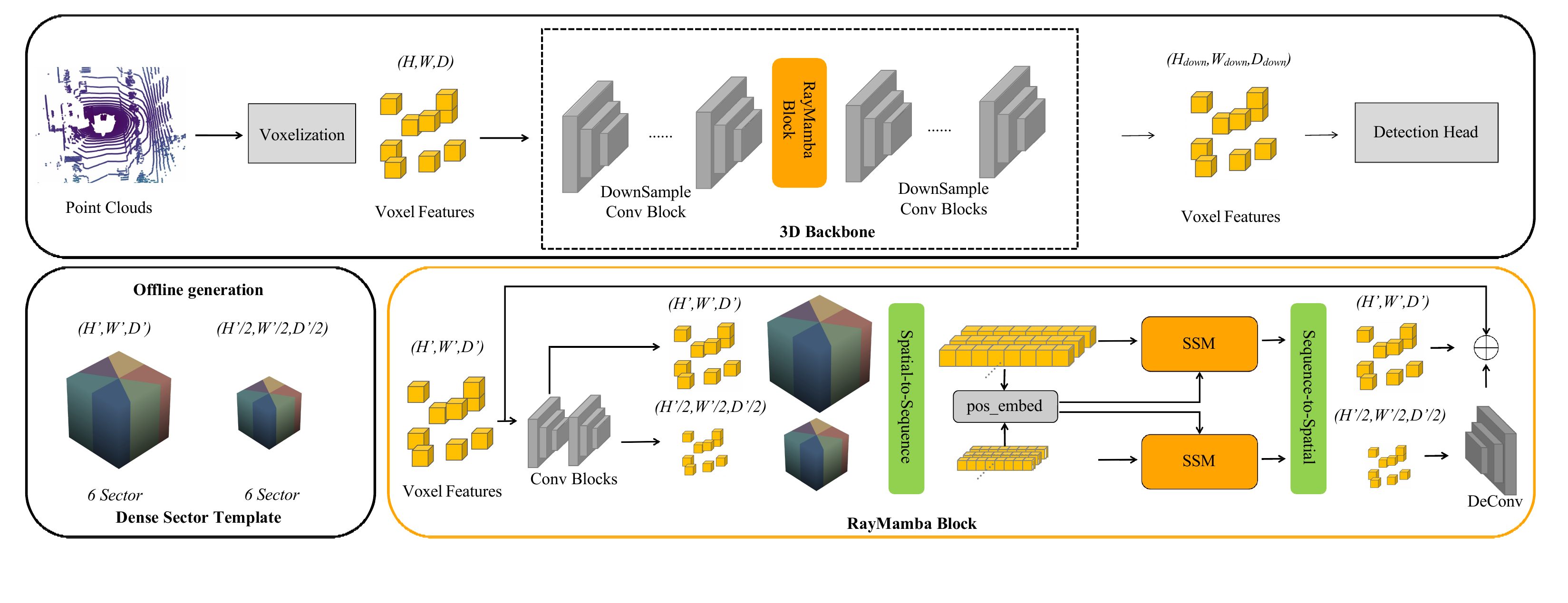}
    \caption{\textbf{Overview of RayMamba.}
    Top: RayMamba blocks are inserted into a sparse 3D convolutional backbone.
    Bottom: Structure of a RayMamba block. RayMamba consists of two components: \textit{Ray-Aligned Serialization}, which converts sparse voxel features into sector-wise ordered sequences using an offline-generated dense sector template, and \textit{SectorMamba3D}, which performs sector-wise sequence modeling before the enhanced features are restored to sparse 3D space through \textit{Sequence-to-Spatial} and sparse deconvolution.}
    \label{fig:main_pipeline}
\end{figure*}

\subsection{Overall Architecture}
Fig.~\ref{fig:main_pipeline} shows the overall pipeline of RayMamba. We design RayMamba as a plug-and-play enhancement for voxel-based 3D detectors and evaluate it on both LiDAR-only and multimodal baselines.

Given an input point cloud, the detector first voxelizes the scene and extracts sparse 3D features with a sparse convolutional backbone, where the voxel resolution is progressively reduced across stages. RayMamba is inserted into this backbone to enhance long-range context modeling for sparse voxels. Since different datasets use different voxel sizes and feature resolutions, RayMamba is applied at corresponding backbone stages under each setting.

Each RayMamba block consists of two components: \textit{Ray-Aligned Serialization} and \textit{SectorMamba3D}. Ray-Aligned Serialization defines a sensor-consistent sector-wise ordering rule and, during runtime, converts active sparse voxels into ordered 1D sequences through template-guided \textit{Spatial-to-Sequence} reordering. Based on these serialized sequences, SectorMamba3D performs sector-wise sequence modeling and restores the enhanced features to sparse 3D space for subsequent fusion with the backbone. The serialization strategy is described in Sec.~\ref{sec:serialization}, while the SectorMamba3D module is introduced in Sec.~\ref{sec:raymamba_block}.

\subsection{Preliminaries: State Space Models and Sequence Modeling}

State Space Models (SSMs) provide an efficient alternative to attention-based sequence modeling. A continuous-time state space model is defined as
\begin{equation}
    h'(t) = \mathbf{A} h(t) + \mathbf{B} x(t),
\end{equation}
\begin{equation}
    y(t) = \mathbf{C} h(t),
\end{equation}
where $h(t)$ is the hidden state and $\mathbf{A}, \mathbf{B}, \mathbf{C}$ are learnable parameters. In practical implementations such as Mamba, this system is discretized for recurrent sequence modeling, leading to linear complexity with respect to sequence length.

To apply Mamba to 3D perception, spatial features must be serialized into 1D sequences before sequence processing. Existing methods typically use generic serialization strategies, such as space-filling curves, local windows, or coordinate sorting. As a result, the quality of sequence modeling depends not only on the Mamba encoder itself, but also on the way spatial features are serialized. This motivates the ray-aligned serialization design introduced next.

\subsection{Ray-Aligned Serialization}
\label{sec:serialization}

\begin{figure*}[t]
    \centering
    \includegraphics[width=\textwidth]{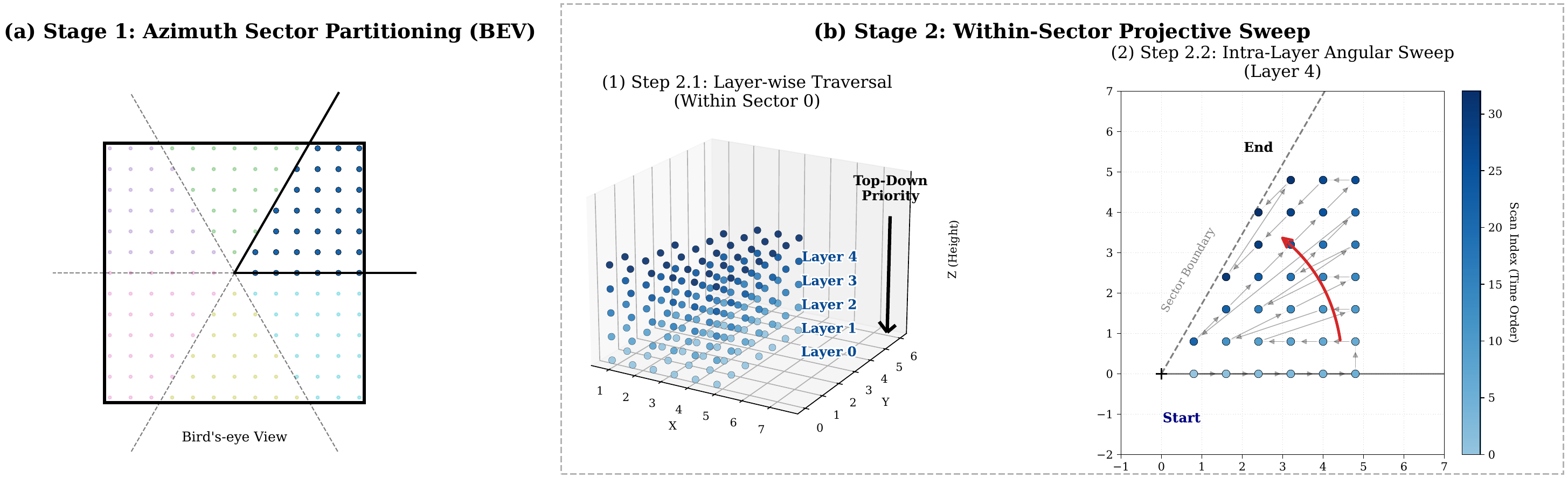}
    \caption{\textbf{Ray-aligned serialization strategy.}
    (a) \textbf{Azimuth sector partitioning:} The BEV space is divided into independent angular sectors to separate directionally distinct regions.
    (b) \textbf{Sector-wise ordering:} Voxels in each sector are serialized by first traversing height layers from top to bottom, introducing a vertical layering prior, and then applying angular ordering within each layer to preserve directional continuity.
    For a fixed voxel grid, the resulting sector assignments and ordering scores are precomputed as a dense sector template, which is queried at runtime to convert active sparse voxels into sector-wise ordered sequences.}
    \label{fig:ray_scan}
\end{figure*}

To better organize sparse voxels for long-range context modeling, we design a ray-aligned serialization mechanism for sparse voxel features. It includes two parts: a sector-wise ordering rule defined on the 3D voxel grid, and a runtime sequence construction process that maps active sparse voxels into ordered 1D sequences. As illustrated in Fig.~\ref{fig:ray_scan}, the ordering rule consists of two hierarchical stages: azimuth sector partitioning, followed by sector-wise ordering. The first stage separates voxels from different directions to avoid mixing unrelated regions in a single global sequence. The second stage is realized through layer-wise traversal and intra-layer angular ordering, which together preserve structured context within each sector.

\paragraph{Azimuth Sector Partitioning}
We first transform voxel coordinates from Cartesian space to polar coordinates centered at the ego vehicle. For a voxel $v_i$ at grid coordinate $(x_i, y_i, z_i)$, its relative BEV coordinate $(r_x, r_y)$ with respect to the center $(c_x, c_y)$ is computed, and its azimuth angle $\theta_i$ is defined as
\begin{equation}
    \theta_i = \left( \operatorname{arctan2}(r_y, r_x) \times \frac{180^\circ}{\pi} + 360^\circ \right) \bmod 360^\circ.
\end{equation}
The full BEV space is then divided into angular sectors with step size $\Delta\theta$, and the sector assignment of voxel $v_i$ is
\begin{equation}
    S(v_i) = \left\lfloor \theta_i / \Delta\theta \right\rfloor .
\end{equation}
In this way, voxels from different azimuth directions are organized into separate sector groups before sequence construction. The sector granularity is controlled by $\Delta\theta$, whose concrete setting is specified in the experiments.

\paragraph{Sector-wise Ordering.}
After sector partitioning, each sector $k$ contains a voxel set $\mathcal{V}_k = \{ v_i \mid S(v_i) = k \}$. We then serialize voxels within each sector using a structured ordering rule.

\textbf{Layer-wise traversal.} Voxels are first organized by descending height, so that higher structures are processed before lower potentially occluded regions. This introduces a vertical layering prior into the sequence.

\textbf{Intra-layer angular ordering.} Within each height layer, voxels are further ordered by azimuth angle to preserve directional continuity inside the sector. Accordingly, for any two voxels $v_i = (x_i, y_i, z_i)$ and $v_j = (x_j, y_j, z_j)$ in the same sector, the ordering relation $v_i \prec v_j$ is defined as
\begin{equation}
    v_i \prec v_j \iff
    \begin{cases}
    z_i > z_j, & \text{if } z_i \neq z_j, \\
    \theta_i < \theta_j, & \text{if } z_i = z_j.
    \end{cases}
\end{equation}

For a fixed voxel grid configuration, the sector assignment and the corresponding ordering score of each discrete voxel location are precomputed offline and stored as a dense sector template. During training and inference, active sparse voxels query this template at their voxel locations, and are reordered through a \textit{Spatial-to-Sequence} operation to form sector-wise ordered sequences. These sequences are then fed into SectorMamba3D for subsequent sequence modeling.

We do not impose an explicit radial-distance order in the final serialization. Instead, the proposed ordering reduces artificial adjacency across unrelated regions while organizing sparse voxels into sector-wise sequences better suited for subsequent Mamba-based modeling.

\begin{table*}[t]
\centering
\caption{Performance comparison on the nuScenes validation set. 'L' denotes LiDAR-only methods, and 'LC' denotes LiDAR-Camera multi-modal methods. The best results within each modality group are highlighted in \textbf{bold}.}
\resizebox{\textwidth}{!}{
\begin{tabular}{l | c | c c | c c c c c c c c c c}
\toprule
Method & Data & mAP & NDS & Car & Truck & C.V & Bus & Trailer & Barrier & Motor. & Bike & Ped. & T.C. \\
\midrule
PointPillars\cite{pointpillars} & L & 34.3 & 49.1 & 79.0 & 38.7 & 5.9 & 52.1 & 25.0 & 42.4 & 21.3 & 0.8 & 60.0 & 18.0 \\
SSN\cite{ssn} & L & 46.7 & 58.2 & 83.0 & 52.1 & 16.2 & 61.7 & 32.1 & 54.1 & 47.9 & 22.3 & 70.1 & 27.2 \\
CenterPoint\cite{centerpoint} & L & 56.4 & 64.7 & 84.6 & 53.9 & 16.5 & 67.0 & 37.5 & \textbf{65.0} & 55.3 & 36.7 & 83.3 & 64.3 \\
+ RayMamba & L & \textbf{57.4} & \textbf{65.6} & \textbf{85.5} & \textbf{54.4} & \textbf{16.7} & \textbf{68.2} & \textbf{38.2} & 63.5 & \textbf{58.7} & \textbf{38.5} & \textbf{84.1} & \textbf{66.6} \\
\midrule
TransFusion\cite{transfusion} & LC & 67.5 & 71.3 & - & - & - & - & - & - & - & - & - & - \\
BEVFusion\cite{bevfusion} & LC & 69.6 & 72.1 & 89.1 & 66.7 & 30.9 & 77.7 & 42.6 & 73.5 & 79.0 & 67.5 & 89.4 & 79.3 \\
BEVFusion4D\cite{bevfusion4d} & LC & 72.0 & 73.5 & \textbf{90.6} & \textbf{70.3} & 32.9 & \textbf{81.5} & 47.1 & 71.6 & 81.5 & 73.0 & 90.2 & 80.9 \\
MV2DFusion\cite{mv2dfusion} & LC & 73.0 & 74.8 & 89.0 & 66.1 & \textbf{37.8} & 80.0 & 45.7 & 75.3 & \textbf{85.2} & 76.6 & \textbf{91.0} & \textbf{83.9} \\
+ RayMamba & LC & \textbf{73.4} & \textbf{75.0} & 90.0 & 67.5 & \textbf{37.8} & 79.2 & \textbf{47.4} & \textbf{75.6} & 84.8 & \textbf{77.1} & 90.5 & 83.6 \\
\bottomrule
\end{tabular}
}
\label{tab:main_results}
\end{table*}

\begin{table}[t]
\centering
\caption{Performance comparison of RayMamba across different scales relative to CenterPoint baseline on nuScenes.}
\label{tab:centerpoint_scales}
\resizebox{\columnwidth}{!}{
\setlength{\tabcolsep}{3pt}
\begin{tabular}{lccccccccc}
\toprule
\multirow{2}{*}{Method} & \multicolumn{2}{c}{Overall} & \multicolumn{2}{c}{0--20m} & \multicolumn{2}{c}{20--40m} & \multicolumn{2}{c}{40--50m} \\
\cmidrule(lr){2-3} \cmidrule(lr){4-5} \cmidrule(lr){6-7} \cmidrule(lr){8-9}
& mAP & NDS & mAP & NDS & mAP & NDS & mAP & NDS \\
\midrule
CenterPoint & 56.4 & 64.7 & 72.3 & 74.0 & 46.0 & 58.6 & 26.7 & 53.4 \\
\midrule
+RayMamba ($11 \times 256^2$) & 57.4 & 65.6 & \textbf{73.2} & \textbf{75.1} & 47.5 & 59.6 & \textbf{29.2} & \textbf{55.0} \\
\textit{vs. baseline} & \textit{+1.0} & \textit{+0.9} & \textit{+0.9} & \textit{+1.1} & \textit{+1.5} & \textit{+1.0} & \textit{+2.5} & \textit{+1.6} \\
\addlinespace
+RayMamba ($21 \times 512^2$) & \textbf{57.7} & \textbf{65.9} & \textbf{73.2} & \textbf{75.1} & \textbf{48.1} & \textbf{60.3} & 27.2 & 54.0 \\
\textit{vs. baseline} & \textit{+1.3} & \textit{+1.2} & \textit{+0.9} & \textit{+1.1} & \textit{+2.1} & \textit{+1.7} & \textit{+0.5} & \textit{+0.6} \\
\addlinespace
+RayMamba (both) & 57.5 & 65.7 & 72.8 & 74.5 & 48.0 & 60.1 & 28.1 & 54.9 \\
\textit{vs. baseline} & \textit{+1.1} & \textit{+1.0} & \textit{+0.5} & \textit{+0.5} & \textit{+2.0} & \textit{+1.5} & \textit{+1.4} & \textit{+1.5} \\
\bottomrule
\end{tabular}
}
\end{table}

\begin{table}[t]
\centering
\caption{Ablation study of the sector angular step $\Delta\theta$ in RayMamba on the MV2DFusion baseline on the nuScenes validation set. Best results are highlighted in bold.}
\label{tab:raymamba_mv2dfusion_angles}
\resizebox{\columnwidth}{!}{
\setlength{\tabcolsep}{3pt}
\begin{tabular}{l cc cc cc cc}
\toprule
\multirow{2}{*}{Method} & \multicolumn{2}{c}{Overall} & \multicolumn{2}{c}{0--20m} & \multicolumn{2}{c}{20--40m} & \multicolumn{2}{c}{40--50m} \\
\cmidrule(lr){2-3} \cmidrule(lr){4-5} \cmidrule(lr){6-7} \cmidrule(lr){8-9}
& mAP & NDS & mAP & NDS & mAP & NDS & mAP & NDS \\
\midrule
MV2DFusion & 73.0 & 74.8 & 82.6 & 80.7 & 66.0 & 70.4 & 43.0 & 63.0 \\
\midrule
+RayMamba ($30^\circ$) & 73.2 & 74.8 & 82.6 & 80.5 & \textbf{66.3} & 70.7 & 43.3 & 63.1 \\
\textit{vs. baseline}  & \textit{+0.1} & \textit{+0.1} & \textit{+0.0} & \textit{-0.2} & \textit{+0.3} & \textit{+0.3} & \textit{+0.3} & \textit{+0.1} \\
\addlinespace
+RayMamba ($60^\circ$) & \textbf{73.4} & \textbf{75.0} & \textbf{83.0} & \textbf{80.9} & \textbf{66.3} & \textbf{70.9} & \textbf{44.6} & \textbf{63.7} \\
\textit{vs. baseline}  & \textit{+0.3} & \textit{+0.3} & \textit{+0.4} & \textit{+0.2} & \textit{+0.3} & \textit{+0.5} & \textit{+1.6} & \textit{+0.7} \\
\addlinespace
+RayMamba ($90^\circ$) & 73.2 & 74.8 & 82.3 & 80.6 & \textbf{66.3} & 70.7 & 44.0 & 63.1 \\
\textit{vs. baseline}  & \textit{+0.1} & \textit{+0.1} & \textit{-0.3} & \textit{-0.1} & \textit{+0.3} & \textit{+0.3} & \textit{+1.0} & \textit{+0.1} \\
\bottomrule
\end{tabular}
}
\end{table}

\subsection{SectorMamba3D}
\label{sec:raymamba_block}

SectorMamba3D serves as the sequence modeling component of RayMamba. Given sparse voxel features from the preceding backbone stage, it first applies lightweight sparse convolutional encoding to produce two feature branches at the original and lower resolutions. This stage also introduces local aggregation, which helps alleviate discontinuities near sector boundaries.

Based on the sector-wise ordered sequences constructed by Ray-Aligned Serialization, SectorMamba3D augments each sequence with learnable positional embeddings and performs Mamba-based sequence modeling independently for each batch-sector pair; empty sectors are skipped. The sector-wise sequence modeling is formulated as
\begin{equation}
    \mathbf{H}_{b,k}^{(l)} = \operatorname{Mamba}^{(l)}\!\left(\tilde{\mathbf{F}}_{b,k}^{(l)}\right),
\end{equation}
where $\tilde{\mathbf{F}}_{b,k}^{(l)}$ denotes the serialized features after positional encoding.

After sequence modeling, the enhanced features are scattered back to their original sparse voxel locations through the inverse indices. The low-resolution branch is further upsampled by sparse inverse convolution and fused with the higher-resolution branch through element-wise addition. In this way, SectorMamba3D enriches sparse voxel features with sector-wise long-range context while remaining fully compatible with standard sparse convolution pipelines.

\subsection{Efficiency and Implementation}
\label{sec:efficiency}

RayMamba introduces only modest overhead. For a fixed voxel grid configuration, the dense sector template is generated offline, storing the sector assignment and ordering score of each discrete voxel location. During training and inference, active sparse voxels query this template at runtime and are reordered through a \textit{Spatial-to-Sequence} operation to form sector-wise ordered sequences. Since this process only requires template lookup and CUDA-based parallel sorting over active voxels, the runtime overhead remains small.

The subsequent sequence modeling stage inherits the linear complexity of state space models, making RayMamba efficient for sparse 3D feature maps where dense global attention is impractical. Detector-specific placement and configuration details are provided in Sec.~\ref{sec:exp_details}.
\begin{figure*}[t]
    \centering
    \includegraphics[width=\textwidth]{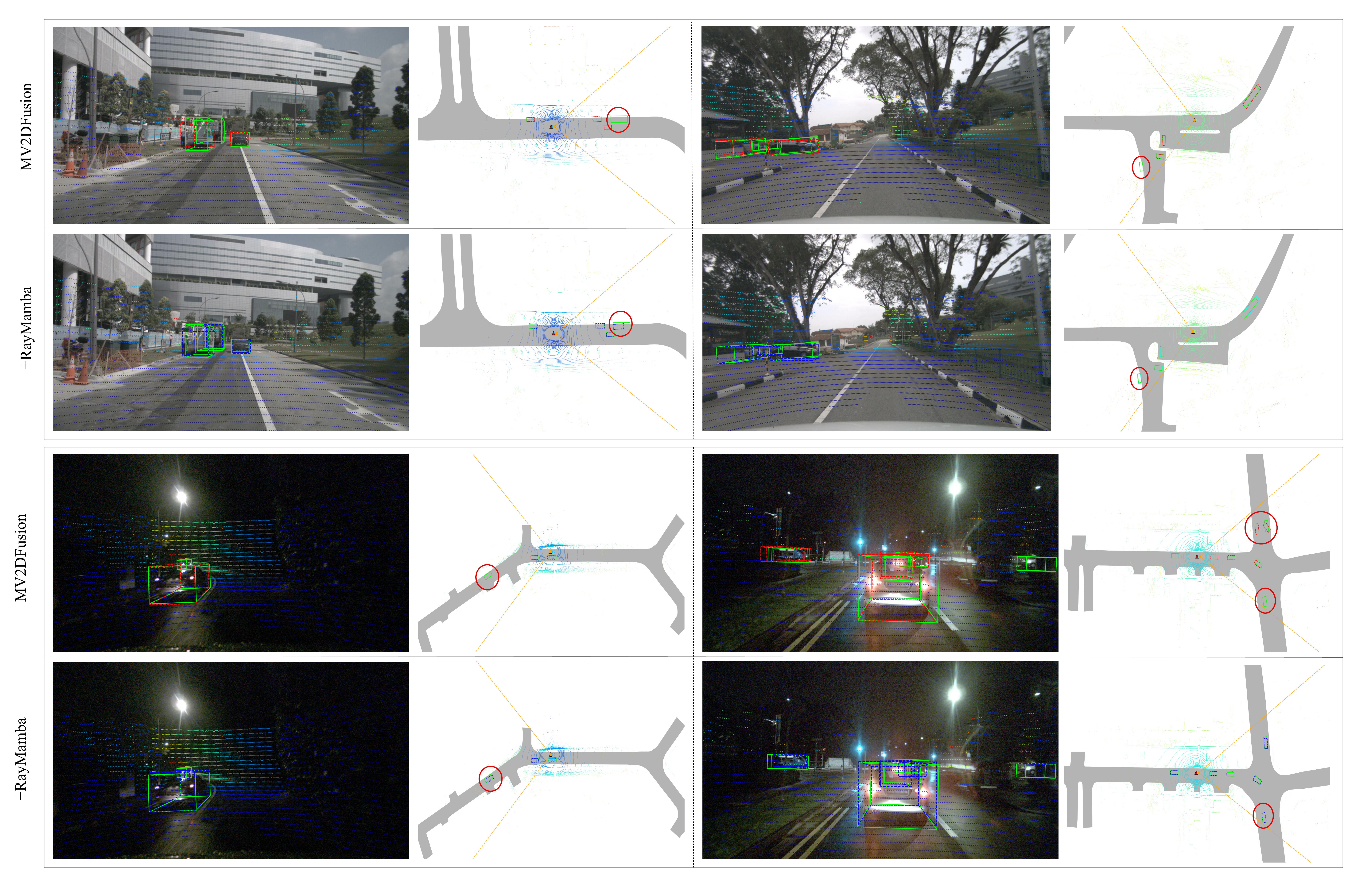}
    \caption{Qualitative comparison on challenging long-range occluded targets. Green boxes denote ground truth, red dashed boxes denote baseline predictions, and blue dashed boxes denote RayMamba predictions.}
    \label{fig:vis_bev}
\end{figure*}

\section{Experiments}
\subsection{Experimental Setup}
\label{sec:exp_details}

We conduct the main experiments on nuScenes~\cite{nuscenes} and use Argoverse 2 (AV2)~\cite{argoverse} for further long-range validation. nuScenes is a widely used autonomous driving benchmark with 1,000 scenes and 10 object categories, evaluated by mAP and NDS. AV2 covers a perception range of up to 200\,m and is therefore more suitable for analyzing far-field detection; we follow the official protocol and report mAP and CDS.

On nuScenes, we evaluate RayMamba on the LiDAR-only baseline CenterPoint and the multimodal baseline MV2DFusion. On AV2, we further validate its generalization on the fully sparse baseline VoxelNeXt. Since the official nuScenes test server was unavailable during our experiments, all nuScenes results are reported on the validation set. In addition to the official overall metrics, we further report distance-based results on 0--20\,m, 20--40\,m, and 40--50\,m.

CenterPoint is trained for 20 epochs with batch size 4, MV2DFusion for 24 epochs with batch size 8, and VoxelNeXt on AV2 for 6 epochs with batch size 16. Unless otherwise specified, all optimization and data augmentation settings follow the corresponding baselines. For CenterPoint, the voxel size is set to [0.1, 0.1, 0.2], with a default RayMamba template size of [11, 256, 256]. For MV2DFusion, we follow the original setting with image resolution 1600 $\times$ 640, voxel size [0.2, 0.2, 0.2], and RayMamba template size [20, 272, 272]. For VoxelNeXt on AV2, the voxel size is [0.1, 0.1, 0.2], and the default template size is [51, 1000, 1000].

Since SectorMamba3D is applied independently to each sector, smaller sector angles lead to more sectors and higher runtime cost. Based on the ablation study, we use $60^\circ$ as the default setting, which provides the best trade-off between accuracy and efficiency. 
\begin{table}[h]
\centering
\caption{Ablation study of explicit distance sorting in RayMamba on the MV2DFusion baseline on the nuScenes validation set. Best results are highlighted in bold.}
\label{tab:ablation_dist_sort}
\resizebox{\columnwidth}{!}{
\begin{tabular}{l cccccccc}
\toprule
\multirow{2}{*}{\textbf{Method}} & \multicolumn{2}{c}{\textbf{Overall}} & \multicolumn{2}{c}{\textbf{0--20m}} & \multicolumn{2}{c}{\textbf{20--40m}} & \multicolumn{2}{c}{\textbf{40--50m}} \\
\cmidrule(lr){2-3} \cmidrule(lr){4-5} \cmidrule(lr){6-7} \cmidrule(lr){8-9}
& mAP & NDS & mAP & NDS & mAP & NDS & mAP & NDS \\
\midrule
MV2DFusion + RayMamba & \textbf{73.4} & \textbf{75.0} & \textbf{83.0} & \textbf{80.9} & \textbf{66.3} & \textbf{70.9} & 44.6 & 63.7 \\
\quad + distance sort & 72.9 & 74.7 & 82.7 & \textbf{80.9} & 65.3 & 70.0 & \textbf{44.7} & \textbf{63.8} \\
\bottomrule
\end{tabular}
}
\end{table}

\begin{table}[t]
\centering
\caption{Efficiency and performance comparison on the nuScenes validation set. All metrics are evaluated in the 40--50\,m range.}
\label{tab:combined_efficiency}
\resizebox{\columnwidth}{!}{%
\begin{tabular}{lcccc}
\toprule
Model & mAP$_{40-50m}$ $\uparrow$ & NDS$_{40-50m}$ $\uparrow$ & Params (M) $\downarrow$ & FPS $\uparrow$ \\
\midrule
\multicolumn{5}{c}{\textbf{Base Architecture: MV2DFusion (Multi-modal)}} \\
\midrule
Baseline & 42.99 & 63.02 & 101.27 & 1.05 \\
+ \textbf{RayMamba} & 44.59 & 63.70 & 103.74 & 1.01 \\
\textbf{Improvement} & \textbf{+1.60} & \textbf{+0.68} & \textbf{+2.47} & \textbf{-0.04} \\
\midrule
\multicolumn{5}{c}{\textbf{Base Architecture: CenterPoint (LiDAR-only)}} \\
\midrule
Baseline & 26.74 & 53.44 & 8.94 & 11.20 \\
+ \textbf{RayMamba} & 29.23 & 55.03 & 9.34 & 9.57 \\
\textbf{Improvement} & \textbf{+2.49} & \textbf{+1.59} & \textbf{+0.40} & \textbf{-1.63} \\
\bottomrule
\end{tabular}%
}
\end{table}

\subsection{Main Results on nuScenes}

\begin{table*}[t]
\centering
\caption{Further validation on the AV2validation set. RayMamba is applied to the VoxelNeXt baseline to examine its generalization in more extreme long-range sparse scenes.}
\label{tab:precision_details_av2}
\resizebox{\textwidth}{!}{
\begin{tabular}{l | ccccccccccccccccccccc}
\toprule
\textbf{Methods} & 
\rotatebox{90}{\textit{Average}} & 
\rotatebox{90}{\textit{Vehicle}} & 
\rotatebox{90}{\textit{Bus}} & 
\rotatebox{90}{\textit{Pedestrian}} & 
\rotatebox{90}{\textit{Stop Sign}} & 
\rotatebox{90}{\textit{Box Truck}} & 
\rotatebox{90}{\textit{Bollard}} & 
\rotatebox{90}{\textit{C-Barrel}} & 
\rotatebox{90}{\textit{Motorcyclist}} & 
\rotatebox{90}{\textit{MPC-Sign}} & 
\rotatebox{90}{\textit{Motorcycle}} & 
\rotatebox{90}{\textit{Bicycle}} & 
\rotatebox{90}{\textit{A-Bus}} & 
\rotatebox{90}{\textit{School Bus}} & 
\rotatebox{90}{\textit{Truck Cab}} & 
\rotatebox{90}{\textit{C-Cone}} & 
\rotatebox{90}{\textit{V-Trailer}} & 
\rotatebox{90}{\textit{Sign}} & 
\rotatebox{90}{\textit{Large Vehicle}} & 
\rotatebox{90}{\textit{Stroller}} & 
\rotatebox{90}{\textit{Bicyclist}} \\
\midrule
CenterPoint\cite{centerpoint}        & 22.0 & 67.6 & 38.9 & 46.5 & 16.9 & 37.4 & 40.1 & 32.2 & 28.6 & 27.4 & 33.4 & 24.5 & 8.7 & 25.8 & \textbf{22.6} & 29.5 & 22.4 & 6.3 & 3.9 & 0.5 & 20.1 \\
FSD\cite{fsd}     & 28.2 & 68.1 & \textbf{40.9} & 59.0 & 29.0 & 38.5 & 41.8 & 42.6 & 39.7 & 26.2 & \textbf{49.0} & 38.6 & 20.4 & \textbf{30.5} & 14.8 & 41.2 & \textbf{26.9} & 11.9 & 5.9 & 13.8 & 33.4 \\
\midrule
VoxelNeXt\cite{voxelnext}          & 30.3 & 72.1 & 38.5 & 63.2 & 37.5 & \textbf{40.5} & 52.8 & 64.9 & 44.4 & \textbf{35.3} & 44.8 & 39.3 & \textbf{21.5} & 27.5 & 17.4 & 45.4 & 21.2 & 16.4 & \textbf{7.1} & 12.3 & 33.3 \\
+ RayMamba         & \textbf{31.2} & \textbf{73.7} & 39.8 & \textbf{64.6} & \textbf{40.1} & \textbf{40.5} & \textbf{56.5} & \textbf{69.3} & \textbf{46.0} & 34.5 & 47.3 & \textbf{40.9} & 20.6 & 26.4 & 19.6 & \textbf{45.5} & 21.5 & \textbf{16.6} & 6.1 & \textbf{15.7} & \textbf{35.0} \\
\bottomrule
\end{tabular}
}
\end{table*}

\begin{table}[t]
\centering
\caption{Ablation study of RayMamba template sizes on the VoxelNeXt baseline on the AV2validation set. Best results are highlighted in bold.}
\label{tab:voxelnext_scale}
\resizebox{\columnwidth}{!}{
\setlength{\tabcolsep}{3pt}
\begin{tabular}{l cc cc cc cc}
\toprule
\multirow{2}{*}{Method} & \multicolumn{2}{c}{Overall} & \multicolumn{2}{c}{0--30m} & \multicolumn{2}{c}{30--50m} & \multicolumn{2}{c}{50--150m} \\
\cmidrule(lr){2-3} \cmidrule(lr){4-5} \cmidrule(lr){6-7} \cmidrule(lr){8-9}
& mAP & CDS & mAP & CDS & mAP & CDS & mAP & CDS \\
\midrule
VoxelNext & 30.3 & 22.9 & 48.7 & 38.5 & 34.2 & 26.3 & 16.5 & 11.7 \\
\midrule
+RayMamba ($13\times250^2$) & 30.2 & 22.8 & 48.0 & 38.1 & 33.3 & 25.6 & 17.3 & 12.3 \\
\textit{vs. baseline} & \textit{-0.1} & \textit{-0.1} & \textit{-0.7} & \textit{-0.4} & \textit{-0.9} & \textit{-0.7} & \textit{+0.8} & \textit{+0.6} \\
\addlinespace
+RayMamba ($26\times500^2$) & 30.6 & 23.1 & 48.2 & 38.2 & \textbf{34.4} & \textbf{26.4} & 17.4 & 12.4 \\
\textit{vs. baseline} & \textit{+0.3} & \textit{+0.2} & \textit{-0.5} & \textit{-0.3} & \textit{+0.2} & \textit{+0.1} & \textit{+0.9} & \textit{+0.7} \\
\addlinespace
+RayMamba ($51\times1000^2$) & \textbf{31.2} & \textbf{23.7} & 49.5 & \textbf{39.3} & 33.7 & 26.0 & \textbf{18.0} & \textbf{12.8} \\
\textit{vs. baseline} & \textit{+0.9} & \textit{+0.8} & \textit{+0.8} & \textit{+0.8} & \textit{-0.5} & \textit{-0.3} & \textit{+1.5} & \textit{+1.1} \\
\addlinespace
+RayMamba ($101\times2000^2$) & 30.8 & 23.3 & \textbf{49.6} & \textbf{39.3} & 34.2 & 26.3 & 16.5 & 11.8 \\
\textit{vs. baseline} & \textit{+0.5} & \textit{+0.4} & \textit{+0.9} & \textit{+0.8} & \textit{\phantom{+}0.0} & \textit{\phantom{+}0.0} & \textit{\phantom{+}0.0} & \textit{+0.1} \\
\bottomrule
\end{tabular}
}
\end{table}

\begin{table}[h]
\centering
\caption{Comparison of sequence serialization strategies on the AV2 validation set based on the VoxelNeXt baseline. RayMamba and HilbertMamba use the same template-size configuration for a fair comparison. Best results are highlighted in bold.}
\label{tab:sorting_comparison}
\resizebox{\columnwidth}{!}{
\begin{tabular}{l cccccccc}
\toprule
\multirow{2}{*}{\textbf{Method}} & \multicolumn{2}{c}{\textbf{Overall}} & \multicolumn{2}{c}{\textbf{0--30m}} & \multicolumn{2}{c}{\textbf{30--50m}} & \multicolumn{2}{c}{\textbf{50--150m}} \\
\cmidrule(lr){2-3} \cmidrule(lr){4-5} \cmidrule(lr){6-7} \cmidrule(lr){8-9}
& mAP & CDS & mAP & CDS & mAP & CDS & mAP & CDS \\
\midrule
VoxelNeXt + RayMamba     & \textbf{31.2} & \textbf{23.7} & \textbf{49.5} & \textbf{39.3} & \textbf{33.7} & \textbf{26.0} & \textbf{18.0} & \textbf{12.8} \\
VoxelNeXt + HilbertMamba & 30.6 & 23.1 & 48.3 & 38.2 & \textbf{33.7} & 25.9 & 17.5 & 12.5 \\
\bottomrule
\end{tabular}
}
\end{table}

Table~\ref{tab:main_results} reports the main results on the nuScenes validation set. RayMamba consistently improves strong baselines in both the LiDAR-only and LiDAR-camera settings, demonstrating its effectiveness and compatibility with different detector architectures.

For LiDAR-only detection, RayMamba boosts CenterPoint from 56.4 to 57.4 mAP and from 64.7 to 65.6 NDS, achieving the best performance among the compared LiDAR-only methods. The gains are particularly notable for sparse and challenging categories such as motorcycle, bicycle, and traffic cone.

For LiDAR-camera detection, adding RayMamba to MV2DFusion further improves performance from 73.0 to 73.4 mAP and from 74.8 to 75.0 NDS. Although the margin is smaller, the consistent gain on a strong multimodal baseline still confirms the effectiveness of enhancing the LiDAR branch with RayMamba.

\subsection{Ablation Studies on nuScenes}

Table~\ref{tab:centerpoint_scales} studies the insertion scale of RayMamba on CenterPoint. All settings improve the baseline, indicating that RayMamba is effective across different feature resolutions. The larger template size $(21 \times 512^2)$ achieves the best overall performance, while the smaller template size $(11 \times 256^2)$ performs best in the 40--50\,m range. Applying RayMamba at both scales does not further improve over the best single-scale setting, suggesting limited benefit from direct multi-scale stacking.

Table~\ref{tab:raymamba_mv2dfusion_angles} studies the sector angular step on MV2DFusion. Since its LiDAR backbone is frozen, RayMamba is applied only to the backbone output, and we therefore focus on the effect of serialization granularity. Among all settings, $\Delta\theta=60^\circ$ achieves the best overall and far-range performance, indicating the best balance between directional separation and contextual continuity. Fig.~\ref{fig:vis_bev} further shows improved detection quality on challenging long-range occluded targets.

\textbf{Ablation on explicit distance sorting in MV2DFusion.}
We further evaluate whether explicit radial-distance sorting benefits RayMamba. As shown in Table~\ref{tab:ablation_dist_sort}, adding distance sorting degrades the overall performance from 73.4/75.0 to 72.9/74.7 in mAP/NDS. Although it brings marginal gains on a few metrics in the 40--50\,m range, the improvement is negligible and comes at the cost of lower overall accuracy. This result supports our final design choice of not using radial-distance sorting.

\subsection{Efficiency Analysis}

Table~\ref{tab:combined_efficiency} shows that the overhead introduced by RayMamba remains well controlled. On MV2DFusion, it increases the parameter count by only 2.47M and reduces FPS by 0.04, while improving mAP$_{40\text{--}50\text{m}}$/NDS$_{40\text{--}50\text{m}}$ by 1.60/0.68. On CenterPoint, RayMamba brings a larger far-range gain of 2.49 mAP and 1.59 NDS with only 0.40M additional parameters. Although the FPS drop is more noticeable on this lightweight baseline, the overall trade-off is still favorable given the clear improvement in the most challenging long-range regime.

\subsection{Further Validation on AV2 with VoxelNeXt}

We further evaluate RayMamba on AV2 with VoxelNeXt to verify its generalization in more extreme long-range sparse scenes. As shown in Table~\ref{tab:precision_details_av2}, RayMamba improves the overall mAP from 30.3 to 31.2. Representative gains can be observed on Vehicle, Pedestrian, Bollard, and Bicyclist, showing that the proposed ray-aligned serialization remains effective under the larger perception range and higher sparsity of AV2.

\textbf{Ablation on insertion scales in VoxelNeXt.}
Table~\ref{tab:voxelnext_scale} shows that the default RayMamba size [51, 1000, 1000] achieves the best overall performance on AV2, improving mAP/CDS from 30.3/22.9 to 31.2/23.7. It also performs best in the farthest 50--150\,m range. In comparison, smaller template sizes are less effective, while a larger size does not bring further gains, indicating that the default setting provides the best trade-off for large-scale sparse scenes.

\textbf{Comparison with Hilbert-based serialization on AV2.}
Table~\ref{tab:sorting_comparison} compares RayMamba with a Hilbert-curve-based variant under the same template-size setting. RayMamba achieves better overall performance (31.2/23.7 vs. 30.6/23.1 in mAP/CDS) and shows clearer advantages in the 0--30\,m and 50--150\,m ranges.

\section{Conclusion}
In this paper, we presented RayMamba, a ray-aligned sequence modeling module for long-range 3D object detection. By replacing generic spatial serialization with a sensor-geometry-aware sector-wise ordering strategy, RayMamba provides more informative context transitions for Mamba-based modeling in sparse far-field scenes. The proposed design is lightweight, compatible with existing sparse convolutional pipelines, and applicable to both LiDAR-only and multimodal detectors. Extensive experiments on nuScenes and AV2show that RayMamba consistently improves overall detection performance, with especially clear gains in long-range regions. These results demonstrate that geometry-aware serialization is an effective direction for enhancing sequence modeling in long-range 3D perception.

\textbf{Limitation and future works}
One limitation of the current design is that the sector-wise Mamba module is executed serially across sectors. As the sector angle decreases, the number of sectors increases, leading to a noticeable drop in inference efficiency. This trade-off becomes more evident on large-scale datasets such as Argoverse 2, where smaller sector angles (e.g., $15^\circ$) can bring additional performance gains, but at a substantially higher runtime cost. In future work, we plan to investigate parallelized sector-wise sequence modeling to better support larger-range and denser-scene perception settings.

\FloatBarrier
\bibliographystyle{IEEEtran} 
\bibliography{main}

\begin{thebibliography}{10}
\providecommand{\url}[1]{#1}
\csname url@samestyle\endcsname
\providecommand{\newblock}{\relax}
\providecommand{\bibinfo}[2]{#2}
\providecommand{\BIBentrySTDinterwordspacing}{\spaceskip=0pt\relax}
\providecommand{\BIBentryALTinterwordstretchfactor}{4}
\providecommand{\BIBentryALTinterwordspacing}{\spaceskip=\fontdimen2\font plus
\BIBentryALTinterwordstretchfactor\fontdimen3\font minus
  \fontdimen4\font\relax}
\providecommand{\BIBforeignlanguage}[2]{{%
\expandafter\ifx\csname l@#1\endcsname\relax
\typeout{** WARNING: IEEEtran.bst: No hyphenation pattern has been}%
\typeout{** loaded for the language `#1'. Using the pattern for}%
\typeout{** the default language instead.}%
\else
\language=\csname l@#1\endcsname
\fi
#2}}
\providecommand{\BIBdecl}{\relax}
\BIBdecl

\bibitem{nuscenes}
H.~Caesar, V.~Bankiti, A.~H. Lang, S.~Vora, V.~E. Liong, Q.~Xu, A.~Krishnan,
  Y.~Pan, G.~Baldan, and O.~Beijbom, ``nuscenes: A multimodal dataset for
  autonomous driving,'' in \emph{Proceedings of the IEEE/CVF conference on
  computer vision and pattern recognition}, 2020, pp. 11\,621--11\,631.

\bibitem{argoverse}
B.~Wilson, W.~Qi, T.~Agarwal, J.~Lambert, J.~Singh, S.~Khandelwal, B.~Pan,
  R.~Kumar, A.~Hartnett, J.~K. Pontes \emph{et~al.}, ``Argoverse 2: Next
  generation datasets for self-driving perception and forecasting,''
  \emph{arXiv preprint arXiv:2301.00493}, 2023.

\bibitem{second}
Y.~Yan, Y.~Mao, and B.~Li, ``Second: Sparsely embedded convolutional
  detection,'' \emph{Sensors}, vol.~18, no.~10, p. 3337, 2018.

\bibitem{voxelnext}
Y.~Chen, J.~Liu, X.~Zhang, X.~Qi, and J.~Jia, ``Voxelnext: Fully sparse
  voxelnet for 3d object detection and tracking,'' in \emph{Proceedings of the
  IEEE/CVF conference on computer vision and pattern recognition}, 2023, pp.
  21\,674--21\,683.

\bibitem{sphericaltransformer}
X.~Lai, Y.~Chen, F.~Lu, J.~Liu, and J.~Jia, ``Spherical transformer for
  lidar-based 3d recognition,'' in \emph{Proceedings of the IEEE/CVF conference
  on computer vision and pattern recognition}, 2023, pp. 17\,545--17\,555.

\bibitem{transformer1}
J.~Mao, Y.~Xue, M.~Niu, H.~Bai, J.~Feng, X.~Liang, H.~Xu, and C.~Xu, ``Voxel
  transformer for 3d object detection,'' in \emph{Proceedings of the IEEE/CVF
  international conference on computer vision}, 2021, pp. 3164--3173.

\bibitem{transformer2}
L.~Fan, Z.~Pang, T.~Zhang, Y.-X. Wang, H.~Zhao, F.~Wang, N.~Wang, and Z.~Zhang,
  ``Embracing single stride 3d object detector with sparse transformer,'' in
  \emph{Proceedings of the IEEE/CVF conference on computer vision and pattern
  recognition}, 2022, pp. 8458--8468.

\bibitem{transformer3}
Z.~Zhou, X.~Zhao, Y.~Wang, P.~Wang, and H.~Foroosh, ``Centerformer:
  Center-based transformer for 3d object detection,'' in \emph{European
  Conference on Computer Vision}.\hskip 1em plus 0.5em minus 0.4em\relax
  Springer, 2022, pp. 496--513.

\bibitem{transformer4}
H.~Wang, C.~Shi, S.~Shi, M.~Lei, S.~Wang, D.~He, B.~Schiele, and L.~Wang,
  ``Dsvt: Dynamic sparse voxel transformer with rotated sets,'' in
  \emph{Proceedings of the IEEE/CVF Conference on Computer Vision and Pattern
  Recognition}, 2023, pp. 13\,520--13\,529.

\bibitem{transformer5}
C.~Zhou, Y.~Zhang, J.~Chen, and D.~Huang, ``Octr: Octree-based transformer for
  3d object detection,'' in \emph{Proceedings of the IEEE/CVF conference on
  computer vision and pattern recognition}, 2023, pp. 5166--5175.

\bibitem{unimamba}
X.~Jin, H.~Su, K.~Liu, C.~Ma, W.~Wu, F.~Hui, and J.~Yan, ``Unimamba: Unified
  spatial-channel representation learning with group-efficient mamba for
  lidar-based 3d object detection,'' in \emph{Proceedings of the Computer
  Vision and Pattern Recognition Conference}, 2025, pp. 1407--1417.

\bibitem{voxelmamba}
G.~Zhang, L.~Fan, C.~He, Z.~Lei, Z.~Zhang, and L.~Zhang, ``Voxel mamba:
  Group-free state space models for point cloud based 3d object detection,''
  \emph{Advances in Neural Information Processing Systems}, vol.~37, pp.
  81\,489--81\,509, 2024.

\bibitem{mambafusion}
H.~Wang, J.~Gao, W.~Hu, and Z.~Zhang, ``Mambafusion: Height-fidelity dense
  global fusion for multi-modal 3d object detection,'' \emph{arXiv preprint
  arXiv:2507.04369}, 2025.

\bibitem{centerpoint}
T.~Yin, X.~Zhou, and P.~Krahenbuhl, ``Center-based 3d object detection and
  tracking,'' in \emph{Proceedings of the IEEE/CVF conference on computer
  vision and pattern recognition}, 2021, pp. 11\,784--11\,793.

\bibitem{mv2dfusion}
Z.~Wang, Z.~Huang, Y.~Gao, N.~Wang, and S.~Liu, ``Mv2dfusion: Leveraging
  modality-specific object semantics for multi-modal 3d detection,'' \emph{IEEE
  Transactions on Pattern Analysis and Machine Intelligence}, 2025.

\bibitem{point1}
Y.~Chen, S.~Liu, X.~Shen, and J.~Jia, ``Fast point r-cnn,'' in
  \emph{Proceedings of the IEEE/CVF international conference on computer
  vision}, 2019, pp. 9775--9784.

\bibitem{point2}
Z.~Yang, Y.~Sun, S.~Liu, X.~Shen, and J.~Jia, ``Std: Sparse-to-dense 3d object
  detector for point cloud,'' in \emph{Proceedings of the IEEE/CVF
  international conference on computer vision}, 2019, pp. 1951--1960.

\bibitem{point3}
C.~R. Qi, O.~Litany, K.~He, and L.~J. Guibas, ``Deep hough voting for 3d object
  detection in point clouds,'' in \emph{proceedings of the IEEE/CVF
  International Conference on Computer Vision}, 2019, pp. 9277--9286.

\bibitem{point4}
B.~Cheng, L.~Sheng, S.~Shi, M.~Yang, and D.~Xu, ``Back-tracing representative
  points for voting-based 3d object detection in point clouds,'' in
  \emph{Proceedings of the IEEE/CVF conference on computer vision and pattern
  recognition}, 2021, pp. 8963--8972.

\bibitem{point5}
Z.~Liu, Z.~Zhang, Y.~Cao, H.~Hu, and X.~Tong, ``Group-free 3d object detection
  via transformers,'' in \emph{Proceedings of the IEEE/CVF international
  conference on computer vision}, 2021, pp. 2949--2958.

\bibitem{voxelnet}
Y.~Zhou and O.~Tuzel, ``Voxelnet: End-to-end learning for point cloud based 3d
  object detection,'' in \emph{Proceedings of the IEEE conference on computer
  vision and pattern recognition}, 2018, pp. 4490--4499.

\bibitem{voxel1}
S.~Dong, L.~Ding, H.~Wang, T.~Xu, X.~Xu, J.~Wang, Z.~Bian, Y.~Wang, and J.~Li,
  ``Mssvt: Mixed-scale sparse voxel transformer for 3d object detection on
  point clouds,'' \emph{Advances in Neural Information Processing Systems},
  vol.~35, pp. 11\,615--11\,628, 2022.

\bibitem{voxel3}
Z.~Liu, X.~Zhao, T.~Huang, R.~Hu, Y.~Zhou, and X.~Bai, ``Tanet: Robust 3d
  object detection from point clouds with triple attention,'' in
  \emph{Proceedings of the AAAI conference on artificial intelligence},
  vol.~34, no.~07, 2020, pp. 11\,677--11\,684.

\bibitem{focalconv}
Y.~Chen, Y.~Li, X.~Zhang, J.~Sun, and J.~Jia, ``Focal sparse convolutional
  networks for 3d object detection,'' in \emph{Proceedings of the IEEE/CVF
  conference on computer vision and pattern recognition}, 2022, pp. 5428--5437.

\bibitem{largekernel3d}
Y.~Chen, J.~Liu, X.~Zhang, X.~Qi, and J.~Jia, ``Largekernel3d: Scaling up
  kernels in 3d sparse cnns,'' in \emph{Proceedings of the IEEE/CVF conference
  on computer vision and pattern recognition}, 2023, pp. 13\,488--13\,498.

\bibitem{link}
T.~Lu, X.~Ding, H.~Liu, G.~Wu, and L.~Wang, ``Link: Linear kernel for
  lidar-based 3d perception,'' in \emph{Proceedings of the IEEE/CVF conference
  on computer vision and pattern Recognition}, 2023, pp. 1105--1115.

\bibitem{fsd}
L.~Fan, F.~Wang, N.~Wang, and Z.-X. Zhang, ``Fully sparse 3d object
  detection,'' \emph{Advances in Neural Information Processing Systems},
  vol.~35, pp. 351--363, 2022.

\bibitem{pointmamba}
D.~Liang, X.~Zhou, W.~Xu, X.~Zhu, Z.~Zou, X.~Ye, X.~Tan, and X.~Bai,
  ``Pointmamba: A simple state space model for point cloud analysis,''
  \emph{Advances in neural information processing systems}, vol.~37, pp.
  32\,653--32\,677, 2024.

\bibitem{lion}
Z.~Liu, J.~Hou, X.~Wang, X.~Ye, J.~Wang, H.~Zhao, and X.~Bai, ``Lion: Linear
  group rnn for 3d object detection in point clouds,'' \emph{Advances in Neural
  Information Processing Systems}, vol.~37, pp. 13\,601--13\,626, 2024.

\bibitem{swformer}
P.~Sun, M.~Tan, W.~Wang, C.~Liu, F.~Xia, Z.~Leng, and D.~Anguelov, ``Swformer:
  Sparse window transformer for 3d object detection in point clouds,'' in
  \emph{European Conference on Computer Vision}.\hskip 1em plus 0.5em minus
  0.4em\relax Springer, 2022, pp. 426--442.

\bibitem{dsvt}
H.~Wang, C.~Shi, S.~Shi, M.~Lei, S.~Wang, D.~He, B.~Schiele, and L.~Wang,
  ``Dsvt: Dynamic sparse voxel transformer with rotated sets,'' in
  \emph{Proceedings of the IEEE/CVF Conference on Computer Vision and Pattern
  Recognition}, 2023, pp. 13\,520--13\,529.

\bibitem{pcm}
T.~Zhang, H.~Yuan, L.~Qi, J.~Zhang, Q.~Zhou, S.~Ji, S.~Yan, and X.~Li, ``Point
  cloud mamba: Point cloud learning via state space model,'' in
  \emph{Proceedings of the AAAI conference on artificial intelligence},
  vol.~39, no.~10, 2025, pp. 10\,121--10\,130.

\bibitem{pointpillars}
A.~H. Lang, S.~Vora, H.~Caesar, L.~Zhou, J.~Yang, and O.~Beijbom,
  ``Pointpillars: Fast encoders for object detection from point clouds,'' in
  \emph{Proceedings of the IEEE/CVF conference on computer vision and pattern
  recognition}, 2019, pp. 12\,697--12\,705.

\bibitem{ssn}
X.~Zhu, Y.~Ma, T.~Wang, Y.~Xu, J.~Shi, and D.~Lin, ``Ssn: Shape signature
  networks for multi-class object detection from point clouds,'' in
  \emph{European Conference on Computer Vision}.\hskip 1em plus 0.5em minus
  0.4em\relax Springer, 2020, pp. 581--597.

\bibitem{transfusion}
X.~Bai, Z.~Hu, X.~Zhu, Q.~Huang, Y.~Chen, H.~Fu, and C.-L. Tai, ``Transfusion:
  Robust lidar-camera fusion for 3d object detection with transformers,'' in
  \emph{Proceedings of the IEEE/CVF conference on computer vision and pattern
  recognition}, 2022, pp. 1090--1099.

\bibitem{bevfusion}
T.~Liang, H.~Xie, K.~Yu, Z.~Xia, Z.~Lin, Y.~Wang, T.~Tang, B.~Wang, and
  Z.~Tang, ``Bevfusion: A simple and robust lidar-camera fusion framework,''
  \emph{Advances in neural information processing systems}, vol.~35, pp.
  10\,421--10\,434, 2022.

\bibitem{bevfusion4d}
H.~Cai, Z.~Zhang, Z.~Zhou, Z.~Li, W.~Ding, and J.~Zhao, ``Bevfusion4d: Learning
  lidar-camera fusion under bird's-eye-view via cross-modality guidance and
  temporal aggregation,'' \emph{arXiv preprint arXiv:2303.17099}, 2023.

\end{thebibliography}
\newpage

\vfill

\end{document}